\documentclass{article}
\usepackage[utf8]{inputenc}
\usepackage[tbtags]{amsmath}
\usepackage{amsfonts,amssymb,mathrsfs,amscd,comment,amsthm}
\usepackage[noend]{algorithm2e}
\usepackage[numbers, compress]{natbib}

\newtheorem{theorem}{Theorem}

\begin{document}
\begin{center}
{\Large Online Algorithm for Aggregating Experts' Predictions with Unbounded Quadratic Loss}

\large\vspace{3mm}
Korotin A., V'yugin V., Burnaev E.\footnote{Skolkovo Institute of Science and Technology. The work is partially supported by the Ministry of Science of Russian Federation, grant 14.756.31.0001}
\end{center}


\small
We consider the problem of online aggregation of experts' predictions with the quadratic loss function. At the beginning of each round $t=1,2,\dots, T$, experts ${n=1,\dots,N}$ provide predictions $\gamma_{t}^{1},\dots,\gamma_{t}^{N}\in \mathbb{H}$ (where $\mathbb{H}$ is a Hilbert space). The player aggregates them to a single prediction $\overline{\gamma_t}\in\mathbb{H}$. Then the nature provides the true outcome $\omega\in\mathbb{H}$. The player and experts $n=1,\dots,N$ suffer losses $h_{t}=\|\omega-\overline{\gamma_{t}}\|^{2}$ and $l_{t}^{n}=\|\omega-\gamma_{t}^{n}\|^{2}$ respectively; the round $t+1$ begins. The goal of the player is to minimize the regret, i.e. the difference between the total loss of the player and the loss of the best expert: $R_{T}=\sum_{t=1}^{T}h_{t}-\min\limits_{n=1,\dots,N}\sum_{t=1}^{T}l_{t}^{n}$.

Online regression is a popular special case of the considered problem, i.e. $\gamma_{t}^{n},\omega_{t}$ are real numbers (the predictions and the outcome) and $\mathbb{H}=\mathbb{R}$ \cite[Section 2.1]{cesa2006prediction}. A more general case is the functional or probabilistic forecasting, for example, $\gamma_{t}^{n},\omega_{t}$ are densities, i.e. elements of $\mathbb{H}=\mathcal{L}^{2}(\mathbb{R}^{D})$, see \cite{korotin2021mixability}.


The problem of online prediction with experts' advice is considered in the game theory \cite{cesa2006prediction} and machine learning \cite{hazan2016introduction}. Existing aggregating algorithms provide strategies which guarantee a constant upper bound on the regret but assume that the losses are bounded. For example, if $l_{t}^{n}\leq B^{2}$ for all $t, n$, the algorithm by \cite[Section 2.1]{cesa2006prediction} guarantees $T$-independent bound ${R_{T}\leq O(B^{2}\ln N)}$. However, the algorithm requires knowing $B$ beforehand.


In this paper, we propose an algorithm for aggregating experts' predictions which does not require a prior knowledge of the upper bound on the losses. The algorithm is based on the exponential reweighing of experts' losses .



\begin{algorithm}[!htb]
\SetAlgorithmName{Algorithm}{empty}{Empty}
\SetKwInOut{Input}{Parameters}
\Input{game length $T$, number of experts $N$, Hilbert space $\mathbb{H}$.}
$B_{0}^{\dagger}\leftarrow 0$; $L_{0}^{n}\leftarrow 0$ for $n=1,2,\dots,N$\;
\For{t=1,2,\dots,T}{
	Experts $n=1,2\dots,N$ provide predictions $\gamma_{t}^{1},\dots,\gamma_{t}^{N}\in \mathbb{H}$\;
	$B_{t}\leftarrow 
	\max\big(B_{t-1}^{\dagger},\max\limits_{n,n'}\|\gamma_{t}^{n}-\gamma_{t}^{n'}\|\big)$; $\eta_{t}\leftarrow \frac{1}{2(B_{t})^{2}}$\;
    $w_{t}^{n}\leftarrow \exp(-\eta_{t}L_{t-1}^{n})/\big[\sum_{n'=1}^{N}\exp(-\eta_{t}L_{t-1}^{n'})\big]$\;
    Player combines the predictions $\overline{\gamma_{t}}\leftarrow\sum_{n=1}^{N}w_{t}^{n}\cdot \gamma_{t}^{n}\in\mathbb{H}$\;
    Nature reveals the true outcome $\omega_{t}\in\mathbb{H}$\;
    Player and experts suffer losses $h_{t}=\|\omega_{t}-\overline{\gamma_{t}}\|^{2}$ and $l_{t}^{n}=\|\omega_{t}-\gamma_{t}^{n}\|^{2}$ \;
    $B_{t}^{\dagger}\leftarrow B_{t}$; $L_{t}^{n}\leftarrow L_{t-1}^{n}+l_{t}^{n}$ for $n=1,2,\dots,N$\;
    \uIf{$\max\limits_{n}(\sqrt{l_{t}^{n}})>B_{t}^{\dagger}$,}{
      	$B_{t}^{\dagger}\leftarrow \sqrt{2}\cdot \max\limits_{n}(\sqrt{l_{t}^{n}})$\;
    }
 }
\caption{player's strategy when the bound on the losses is not known beforehand.}
\label{algorithm-unbounded}
\end{algorithm}
The proposed algorithm assigns weights to the experts proportionally to the inverse exponent of their cumulative losses from the previous steps. The weights are used to perform linear (convex) aggregation of predictions. The learning rate (parameter $\eta$ of the algorithm) changes dynamically allowing the algorithm to adapt to the max observed loss.

\begin{theorem}
The regret of Algorithm \ref{algorithm-unbounded} satisfies $R_{T}\leq O\big(\max\limits_{t,n}l_{t}^{n}\cdot (\ln N +1)\big)$.
\end{theorem}
\vspace{-5mm}\begin{proof}
We consider the step $t$. Let $\mathbb{H}_{t}=\text{Span}\{\omega_{t},\gamma_{t},\dots,\gamma_{t}^{N}\}$. It is a $\leq N+1$-dimensional linear subspace of $\mathbb{H}$. Denote $S_{t}=\bigcap_{n=1}^{N}\{ \gamma\in \mathbb{H}_{t}\mid \|\gamma-\gamma_{t}^{n}\|\leq B_{t}\}$, i.e. the convex set of $\gamma\in \mathbb{H}_{t}$ which are $B_{t}$-close to all the predictions $\gamma_{1},\dots,\gamma_{N}$. Let $\Gamma_{t}=\text{ConvexHull}\{\gamma_{t}^{1},\dots,\gamma_{t}^{N}\}$. Note that $\Gamma_{t}\subset S_{t}$. This follows from $B_{t}\geq \max\limits_{n,n'}\|\gamma_{t}^{n}-\gamma_{t}^{n'}\|$ due to the definition of $B_{t}$ in Algorithm \ref{algorithm-unbounded}.

Let $\lfloor\omega_{t}\rfloor$ be the projection of $\omega_{t}$ to $S_{t}$. We define $f_{t}:\Gamma_{t}\rightarrow \mathbb{R}$ by ${f_{t}(\gamma)=\|\gamma-\lfloor\omega_{t}\rfloor\|^{2}}$. For all $\gamma\in\Gamma_{t}$ it holds true that $\|\gamma-\lfloor\omega_{t}\rfloor\|\leq B_{t}$. By using \cite[Lemma 4.2]{hazan2016introduction} we conclude that $f_{t}$ is $\eta_{t}= \frac{1}{2(B_{t})^{2}}$-exponentially concave function. \hspace{-.3mm}Thus,\hspace{-.3mm} $\exp(-\eta_{t}f(\overline{\gamma_{t}}))\geq \sum_{n=1}^{n}w_{t}^{n}\exp(-\eta_{t}f(\gamma_{t}^{n}))$, where $\overline{\gamma}_{t}$ is the aggregated prediction of the player. We denote $f_{t}(\overline{\gamma}_{t})=\|\overline{\gamma_{t}}-\lfloor\omega_{t}\rfloor\|^{2}$ by $\lfloor h_{t}\rfloor$ and $f(\gamma^{n}_{t})=\|\gamma_{t}^{n}-\lfloor\omega_{t}\rfloor\|^{2}$ by $\lfloor l_{t}^{n}\rfloor$, and obtain the inequality $\exp(-\eta_{t}\lfloor h_{t}\rfloor)\geq \sum_{n=1}^{N}w_{t}^{n}\exp(-\eta_{t}\lfloor l_{t}^{n}\rfloor)$. 

For $\eta>0$ we define $\lfloor m_{t}\rfloor(\eta)=-\frac{1}{\eta}\ln \sum_{n=1}^{N}w_{t}^{n}\exp(-\eta \lfloor l_{t}^{n}\rfloor)$. To begin with, from the previous paragraph it follows that $\lfloor h_{t}\rfloor\leq \lfloor m_{t}\rfloor(\eta_{t})$. Next, for $\eta>0$ we set $m_{t}(\eta)\!=\!-\frac{1}{\eta}\ln \sum_{n=1}^{N}w_{t}^{n}\exp(-\eta l_{t}^{n})$. Since $\lfloor \omega_{t}\rfloor$ is the projection of $\omega_{t}$ to the convex set $S_{t}$, it holds true that $\lfloor l_{t}^{n}\rfloor=\| \gamma_{t}^{n}-\lfloor\omega_{t}\rfloor\|^{2}\leq \| \gamma_{t}^{n}-\omega_{t}\|^{2}=l_{t}^{n}$. Thus, $\forall \eta>0$ we have $m_{t}(\eta)\geq \lfloor m_{t}\rfloor(\eta)$. In particular, for $\eta=\eta_{t}$ we have $\lfloor h_{t}\rfloor\leq \lfloor m_{t}\rfloor(\eta_{t})\leq m_{t}(\eta_{t})$.

Let us prove that $h_{t}\leq \lfloor h_{t}\rfloor + (B_{t}^{\dagger})^2-(B_{t})^2$. If $\omega_{t}\in S_{t}$, then $\omega_{t}=\lfloor \omega_{t}\rfloor$, $h_{t}= \lfloor h_{t}\rfloor$ and $B^{\dagger}_{t}= B_{t}$, which results in the desired inequality. If $\omega_{t}\notin S_{t}$, then $\max\limits_{n}(\sqrt{l_{t}^{n}})>B_{t}$ and $(B_{t}^{\dagger})^{2}= 2\cdot \max\limits_{n}(l_{t}^{n})>\max\limits_{n}(l_{t}^{n})+(B_{t})^{2}$ by the definition in Algorithm 1. Thus, $\lfloor h_{t}\rfloor + (B_{t}^{\dagger})^2-(B_{t})^2> \lfloor h_{t}\rfloor + \max\limits_{n}(l_{t}^{n})\geq \max\limits_{n}(l_{t}^{n})\geq h_{t}$, where the last inequality follows from the convexity of the square function $f_{t}$ and $\overline{\gamma}_{t}\in \Gamma_{t}$.

We combine the derived inequalities and conclude $h_{t}\leq m_{t}(\eta_{t})+ (B_{t}^{\dagger})^2-(B_{t})^2$. We sum  the obtained inequality for $t=1,\dots,T$ and obtain $\sum_{t=1}^{T}h_{t}\leq (B_{T}^{\dagger})^{2}+\sum_{t=1}^{T}m_{t}(\eta_{t})$. Note that $\eta_{1}\geq\eta_{2}\geq\dots\geq \eta_{T}$ is a non-increasing dynamic learning rate. We use \cite[Lemma 2]{de2014follow} to obtain $\sum_{t=1}^{T}m_{t}(\eta_{t})\leq -\frac{1}{\eta_{T}}\ln\sum_{t=1}^{T}\frac{1}{N}\exp(-\eta_{T}L_{T}^{n})$.

The latter quantity does not exceed $-\frac{1}{\eta_{T}}\ln\big(\frac{1}{N}\exp(-\eta_{T}\min\limits_{n}L_{T}^{n})\big)=\frac{\ln N}{\eta_{T}}+\min\limits_{n}L_{T}^{n}$. We immediately obtain the regret bound for our algorithm: $\sum_{t=1}^{T}h_{t}-\min\limits_{n}L_{T}^{n}\leq \frac{\ln N}{\eta_{T}}+(B_{T}^{\dagger})^{2}=2(B_{t})^{2}\cdot \ln N+(B^{\dagger}_{t})^{2}\leq (2\ln N + 1)\cdot (B^{\dagger}_{t})^{2}$. Note that $B_{t}^{\dagger}\leq \max\big(\max\limits_{t, n}\|\gamma_{t}^{n}-\gamma_{t}^{n'}\|,\sqrt{2}\max\limits_{t,n}\sqrt{l_{t}^{n}}\big)$. Due to the triangle inequality, $\forall t,n,n'$ we have $\|\gamma_{t}^{n}-\gamma_{t}^{n'}\|\leq \|\gamma_{t}^{n}-\omega_{t}\|+\|\gamma_{t}^{n'}-\omega_{t}\|=\sqrt{l_{t}^{n}}+\sqrt{l_{t}^{n'}}\leq 2\cdot \max_{n}\sqrt{l_{t}^{n}}$. Thus, $B_{T}^{\dagger}\leq 2\max\limits_{n,t}\sqrt{l_{t}^{n}}$, and the final regret bound is $\sum_{t=1}^{T}h_{t}-\min\limits_{n}L_{T}^{n}\leq 4(2\ln N + 1) \max\limits_{t,n}l_{t}^{n}=O\big(\max\limits_{t,n}l_{t}^{n}\cdot (\ln N +1)\big)$.
\end{proof}
\vspace{-1em}
\bibliographystyle{plain}
\bibliography{references}

\begin{thebibliography}{1}

\bibitem{cesa2006prediction}
Nicolo Cesa-Bianchi and G{\'a}bor Lugosi.
\newblock {\em Prediction, learning, and games}.
\newblock Cambridge university press, 2006.

\bibitem{de2014follow}
Steven De~Rooij, Tim Van~Erven, Peter~D Gr{\"u}nwald, and Wouter~M Koolen.
\newblock Follow the leader if you can, hedge if you must.
\newblock {\em The Journal of Machine Learning Research}, 15(1):1281--1316,
  2014.

\bibitem{hazan2016introduction}
Elad Hazan.
\newblock Introduction to online convex optimization.
\newblock {\em Foundations and Trends in Optimization}, 2(3-4):157--325, 2016.

\bibitem{korotin2021mixability}
Alexander Korotin, Vladimir V’yugin, and Evgeny Burnaev.
\newblock Mixability of integral losses: A key to efficient online aggregation
  of functional and probabilistic forecasts.
\newblock {\em Pattern Recognition}, 120:108175, 2021.

\end{thebibliography}

\end{document}